\documentclass[10pt,twocolumn,letterpaper]{article}

\usepackage{iccv}
\usepackage{times}
\usepackage{epsfig}

\usepackage{graphicx}
\usepackage{float}
\usepackage{amsmath}
\usepackage{amssymb}
\usepackage{pifont}
\usepackage{setspace}
\usepackage{booktabs}
\usepackage{tablefootnote}
\usepackage{subfigure}
\usepackage{multirow}
\usepackage[accsupp]{axessibility}  
\newcommand{\cmark}{\ding{51}}
\newcommand{\xmark}{\ding{55}}

\usepackage[pagebackref=true,breaklinks=true,letterpaper=true,colorlinks,bookmarks=false]{hyperref}
\usepackage[capitalize]{cleveref}
\iccvfinalcopy 
\crefname{section}{Sec.}{Secs.}
\Crefname{section}{Section}{Sections}

\def\iccvPaperID{6728} 

\ificcvfinal\fi

\begin{document}

\title{OFVL-MS: Once for Visual Localization across Multiple Indoor Scenes}

\author{Tao Xie$^{1}$,\ \ Kun Dai$^{1}$,\ \ Siyi Lu$^{1,2}$,\ \  Ke Wang$^{1,}$\footnotemark[1],\ \ Zhiqiang Jiang$^{1}$,\ \  Jinghan Gao$^{1}$,\ \  \\ Dedong Liu$^{1}$,\ \  Jie Xu$^{1}$,\ \ Lijun Zhao$^{1,3,}$\footnotemark[1], Ruifeng Li$^{1,}$\footnotemark[1] \\
\normalsize $^1$Harbin Institute of Technology\ \ \ 
\ 
$^2$China Coal Science and Technology Intelligent Storage Technology Co., Ltd. \\
\normalsize $^3$Harbin Institute of Technology, Zhengzhou Research Institute \\
{\tt\small $\{$xietao1997, wangke, jeff$\_$xu, zhaolj, lrf100$\}$@hit.edu.cn} \\ 
{\tt\small \{20s108237, 19S108222, 20s108237, 22s108222, 22S108237\}@stu.hit.edu.cn }
}

\maketitle
\renewcommand{\thefootnote}{\fnsymbol{footnote}}
\footnotetext[1]{Corresponding author.}
\ificcvfinal\thispagestyle{empty}\fi

\begin{abstract}
	In this work, we seek to predict camera poses across scenes with a multi-task learning manner, where we view the localization of each scene as a new task. 
	We propose OFVL-MS, a unified framework that dispenses with the traditional practice of training a model for each individual scene and relieves gradient conflict induced by optimizing multiple scenes collectively, enabling efficient storage yet precise visual localization for all scenes. 
	Technically, in the forward pass of OFVL-MS, we design a layer-adaptive sharing policy with a learnable score for each layer to automatically determine whether the layer is shared or not. 
    Such sharing policy empowers us to acquire task-shared parameters for a reduction of storage cost  and task-specific parameters for learning scene-related features to alleviate gradient conflict. 
	In the backward pass of OFVL-MS, we introduce a gradient normalization algorithm that homogenizes the gradient magnitude of the task-shared parameters so that all tasks converge at the same pace.
	Furthermore, a sparse penalty loss is applied on the learnable scores to facilitate parameter sharing for all tasks without performance degradation.
	We conduct comprehensive experiments on multiple benchmarks and our new released indoor dataset LIVL, showing that OFVL-MS families significantly outperform the state-of-the-arts with fewer parameters. 
    We also verify that OFVL-MS can generalize to a new scene with much few parameters while gaining superior localization performance.  
    The dataset and evaluation code is available at \href{https://github.com/mooncake199809/UFVL-Net}{https://github.com/mooncake199809/UFVL-Net}.
\end{abstract}

\section{Introduction}
\label{sec:intro}
Visual localization, a challenging task that aims to forecast 6-DOF camera pose on a provided RGB image, is an integral part of several computer vision tasks, such as simultaneous localization and mapping~\cite{wang2021continual, mur2017orb, campos2021orb} and structure-from-motion \cite{ding2019camnet, moran2021deep}. 

Typically, classical structure-based visual localization frameworks~\cite{sattler2016efficient, sarlin2019coarse, yu2020learning, yang2022scenesqueezer} construct 2D keypoints and 3D scene coordinates associations by matching local descriptors, and afterwards use a RANSAC-based PnP algorithm~\cite{fischler1981random, lepetit2009epnp} to retrieve camera pose. 
Recently, with the advancements of deep learning~\cite{xie2023poly, xie2023farp, song2023vp, wang2023multi, wang2023sat}, 
scene coordinate regression (SCoRe) based methods \cite{li2020hierarchical, dosovitskiy2020image, zhou2020kfnet, xie2022deep, guan2021scene, dai2023eaainet}, which trains a convolutional neural network (CNN) to regress the 3D scene coordinate corresponding to each pixel in the input image and calculates camera pose with  PnP algorithm \cite{lepetit2009epnp}, establish state-of-the-art localization performance in small static scenes. 
Compared with structure-based methods, these methods require no database of images or local descriptors and can benefit from high-precision sensors. 
While SCoRe based methods achieve impressive results, they come with some drawbacks. 
Scene coordinate regression is scene-specific and required to be trained for new scenes, resulting in a linear increase in total model size with the number of scenes. 
After witnessing the success of SCoRe-based methods, a naive problem arise: could a single SCoRe-based model predict 3D coordinates for multiple scenes concurrently and generalize to a new scene? 
Solving this problem is a key step towards truly SCoRe-based model deployment on autonomous robots. 

A naive solution to this problem is that using a shared backbone to extract features from multiple scenes and then leveraging different regression heads to regress scene coordinates for each scene. 
Nevertheless, jointly optimizing cross-scene localization with a fully shared backbone exists an insurmountable obstacle, i.e., gradient conflict induced by competition among different tasks for shared parameters, resulting in inferior performance compared with learning tasks separately \cite{liu2021conflict, chen2020just, fifty2021efficiently, guo2018dynamic}. 
Towards this end, we propose OFVL-MS, a unified SCoRe-based framework that optimizes visual localization of multiple scenes collectively. 
OFVL-MS is a multi-task learning (MTL)~\cite{doersch2017multi, kendall2018multi, liu2019end, chennupati2019multinet++, xie2023poly, wang2023mdl, radwan2018vlocnet++, wallingford2022task} framework where localization of each scene is treated as an individual task. 
OFVL-MS offers benefits in terms of model complexity and learning efficiency since substantial parameters of the network are shared among multiple scenes, which renders the model more pragmatic to be deployed on robotics. 
Technically, OFVL-MS eliminates gradient conflict from forward and backward pass. 

In the forward pass, we design a layer-adaptive sharing policy to automatically determine whether each active layer of the backbone is shared or not, from which we derive task-shared parameters for efficient storage and task-specific parameters for mitigating gradient conflict. 
The central idea of the layer-adaptive sharing policy is to transform the layer selection of the backbone into a learnable problem, so that deciding which layers of the backbone to be shared or not can be done during training by solving a joint optimization problem. 
In the backward pass, inspired by gradient homogenization algorithms in classical multi-task learning \cite{javaloy2021rotograd, liu2021towards}, we introduce a gradient normalization algorithm that homogenizes the gradient magnitude of the task-shared parameters across scenes to ensure all tasks converge at a similar but optimal pace, further relieving gradient conflict. 
We also apply a penalty loss on the active layers to prompt all tasks to share as many parameters as possible while improving the performance of some tasks that benefit from the shared parameters, as illustrated in \cref{joint_separate} and \cref{section4_7}. 
Experiments show that OFVL-MS achieves excellent localization performance on several benchmarks, including 7-Scenes dataset\cite{shotton2013scene}, 12-Scenes datasets \cite{valentin2016learning} and our \textbf{released large indoor dataset LIVL} in terms of median positional and rotational errors, etc. 
We also demonstrate that OFVL-MS can generalize to a new scene with much few parameters while maintaining exceptional performance. 

To summarize, the contributions of this work are as follows: (1) We propose OFVL-MS, a unified visual localization framework that optimizes localization tasks of different scenes collectively in a multi-task learning manner. (2) We propose a layer-adaptive sharing policy for OFVL-MS to automatically determine, rather than manually, whether each active layer of backbone is shared or not. A penalty loss is also applied to promote layer sharing across scenes. (3) We introduce a gradient normalization algorithm to homogenize gradient magnitudes of the task-shared parameters, enabling all tasks to converge at same pace. 
(4) We publish a \textbf{new large indoor dataset LIVL} that provides a new test benchmark for visual localization. 
(5) We demonstrate that OFVL-MS can generalize to a new scene with much fewer parameters while retaining superior localization performance. 

%

\section{Related Work}
\textbf{Structured-based Visual Localization}.
The structure-based methodologies \cite{sattler2016efficient, sarlin2019coarse, yu2020learning, yang2022scenesqueezer} utilize local descriptors to establish 2D pixel positions and 3D scene coordinate matches for a given query image, afterwards using a PnP algorithm to recover camera pose.  
However, as opposed to directly matching within an exhaustive 3D map as in \cite{sattler2016efficient}, current state-of-the-art methods \cite{sarlin2019coarse, yu2020learning, yang2022scenesqueezer} employ image retrieval \cite{arandjelovic2016netvlad} to narrow down the searching space and utilize advanced feature matching techniques such as Patch2pix \cite{zhou2021patch2pix}, SuperGlue \cite{sarlin2020superglue}, LoFTR \cite{sun2021loftr}, MatchFormer \cite{wang2022matchformer}, OAMatcher~\cite{dai2023oamatcher}, and DeepMatcher~\cite{xie2023deepmatcher} to generate precise 2D-2D correspondences, which are subsequently elevated to 2D-3D matches. 
The structured-based methods demonstrate state-of-the-art performance in large-scale scenes thanks to expeditious image retrieval techniques and feature matching algorithms, while they are limited in small-scale static scenes such as indoor scenes \cite{li2020hierarchical, huang2021vs}. 
Moreover, in lifelong localization scenarios, the size of the image and feature database increases over time due to the continuous addition of new data. As a result, the memory requirements for on-device localization in VR/AR systems may exceed the available limits. 


\textbf{Learning-based Visual Localization}.
Current learning-based visual localization approaches can be classified into absolute pose regression (APR) \cite{kendall2015posenet, xie2020visual, kendall2016modelling}, relative pose regression (RPR) \cite{abouelnaga2021distillpose, ding2019camnet, turkoglu2021visual}, and scene coordinate regression (SCoRe) \cite{li2020hierarchical, dosovitskiy2020image, zhou2020kfnet, xie2022deep, guan2021scene}. 
The APR methods directly forecast the camera pose via a provided RGB image in an end-to-end way. 
However, such methods can not realize accurate visual localization as they are essentially analogous to approximate pose estimation via image retrieval \cite{walch2017image}.
The RPR methods utilize a neural network to identify the relative pose among the requested image and the most identical image retrieved from the database, which, however, is time-consuming and restricts their practical application. 
The SCoRe approaches directly forecast the 3D scene coordinates, walked by the RANSAC-based PnP algorithm The SCoRe approaches directly forecast the scene coordinates, succeeded by the PnP algorithm to compute camera pose. 
While these methods can be optimized end-to-end and achieve impressive results, they suffer from some drawbacks. 
Pose regression and scene coordinate regression are both scene-specific and must be retrained for new scenes, culminating in a linear increase in total model size with the number of scenes. 
\begin{figure*}[htbp]
	\centering
	\includegraphics[width=0.98\hsize]{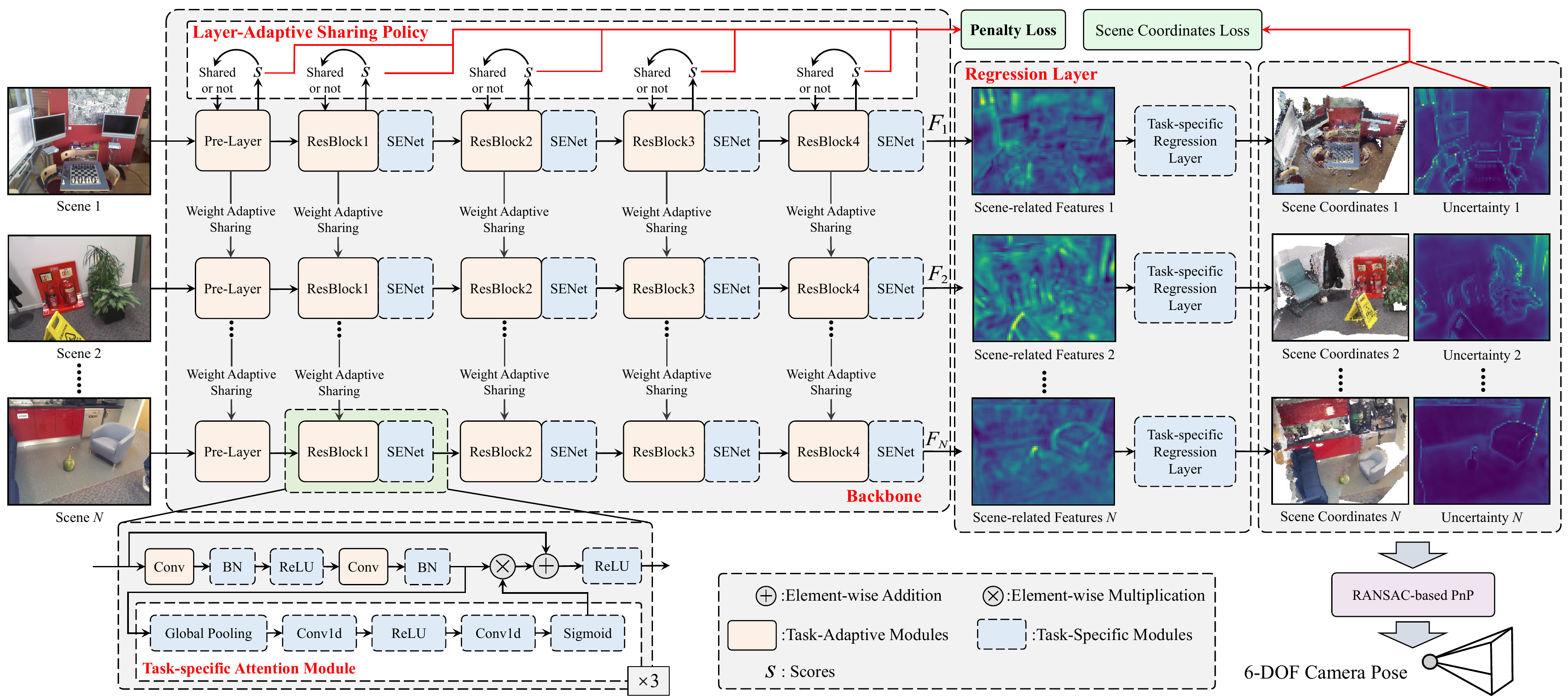}
	\caption{\textbf{Overall of OFVL-MS (using ResNet34~\cite{he2016deep} as backbone).} 
		OFVL-MS jointly optimizes visual localization across scenes and consists of two components, that is, backbone and regression layer.
		The layer-adaptive sharing policy and task-specific attention module are utilized to generate more scene-related features, which are fed into regression layers to predict scene coordinates with uncertainty.
		Besides, the penalty loss is proposed to facilitate OFVL-MS to share parameters as many as possible, realizing efficient storage deployment.
	}
	\label{overall}
\end{figure*}
\begin{figure}[]
	\centering
	\includegraphics[width=0.9\hsize]{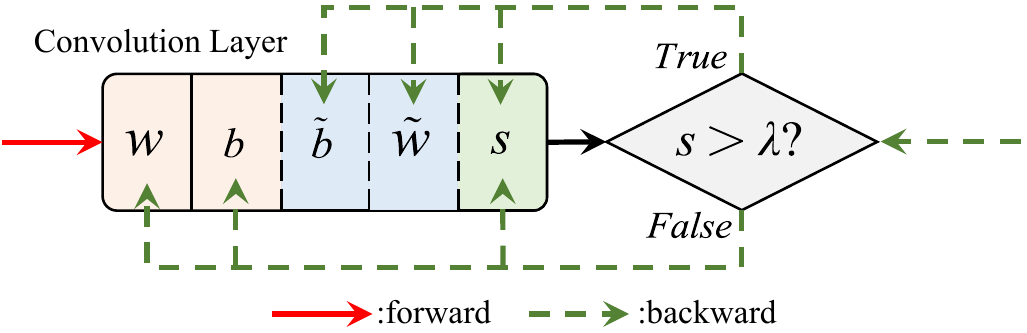}
	\caption{\textbf{Layer-adaptive Sharing Policy.}
		The scores $s$ are utilized to determine which parameters (($w, b, s$) or ($\tilde{w}, \tilde{b}, s$)) to be optimized in current iteration.
	}
	\label{layer-sharing}
\end{figure}

\textbf{Gradient Homogenization over Multi-task Learning (MTL)}.
During the training process of multi-task learning (MTL), the gradient magnitudes and directions of different tasks interact complicatedly together via backpropagation, a phenomenon known as task interference. 
Previous methods \cite{sener2018multi, liu2021towards, chen2018gradnorm, javaloy2021rotograd, maninis2019attentive, sinha2018gradient}  simplify the matter to two categories of gradient discrepancies (i.e., magnitudes and directions of task gradients) and suggest various techniques to reconcile this difference. 
For gradient magnitudes, Sener et al. \cite{sener2018multi} characterize multi-task learning as multi-objective optimization and provide an upper bound for the multi-objective loss. 
Javaloy et al. \cite{javaloy2021rotograd} homogenize the gradient magnitudes through normalizing and scaling, ensuring training convergence.
For gradient direction, Sinha et al. \cite{sinha2018gradient} and Maninis et al. \cite{maninis2019attentive} propose to enable task gradients statistically to be indistinguishable through adversarial training.

\section{Method}
Given a RGB image, the task of visual localization seeks to estimate the rigid transformation $T \in SE(3)$ from camera coordinate system to world coordinate system. 
Such transformation is composed of a 3D rotation matrix $R \in SO(3)$ and a translation vector $t \in \mathbb{R}^{3}$.  

\subsection{Overall}
We propose OFVL-MS, a unified framework that jointly optimizes localization tasks of different scenes in a multi-task learning manner, where we view the visual localization of each scene as a new task. 
OFVL-MS is a two-stage pipeline with scene coordinates prediction followed by a RANSAC-based PnP algorithm to calculate the camera pose $T$. 
Specifically, OFVL-MS takes $N$ RGB images $I_{n} \in \mathbb{R}^{3 \times H \times W}, \ n\in\{1,2,...,N\}$ from different scenes as input and predicts dense 3D scene coordinates $\hat{D_n} = \{\hat{d}_{n,i} = (\hat{x}_{n,i}, \hat{y}_{n,i}, \hat{z}_{n,i}) | i=1,2,3,...,Q \}$ with 1D uncertainty $\hat{U_n} = \{\hat{u}_{n,i} | i=1,2,3,...,Q \}$, where $Q$ is the predicted 3D scene coordinate numbers. 
Thus, we derive $Q$ correspondences between 2D pixel coordinates and 3D scene coordinates. 
Finally, OFVL-MS utilizes RANSAC-based PnP algorithm to calculate 6-DOF camera pose $T_{n}=[R_{n}|t_{n}]$.
In this work, we focus on designing and optimising OFVL-MS, which encourages all tasks to share as many parameters as possible for efficient storage deployment while maintaining superior performance for all tasks.

\subsection{Design OFVL-MS}
As shown in \cref{overall}, OFVL-MS is characterized by two components: backbone and regression layers. 

\textbf{Backbone.} 
The backbone first utilizes a pre-layer with stride of $2$ to map the input image to a higher dimension and lower resolution, and then leverages four ResBlocks \cite{he2016deep} with stride of $(1,1,2,2)$ and several attention modules to extract features. 
The backbone concludes a set of task-shared parameters $\phi^{sh}$ for $N$ tasks and task-specific parameters $\phi^{sp}_{n}$ for task $n$ to transform each input $I_n$ into an intermediate representation $F_n =  f(I_n; \phi^{sh}, \phi^{sp}_{n}) \in \mathbb{R}^{C_{o} \times H_{o} \times W_{o}}$, where $C_{o}$ is the dimension of $F_n$, $H_{0}=H/8$, $W_{0}=W/8$. 

\textbf{Regression Layer.} 
Additionally, each task $n$ has a regression layer $h$, with exclusive parameters $\theta_n$, which takes $F_n$ as input and predicts 3D scene coordinate $\hat{D_n}$ as well as 1D uncertainty $\hat{U_n}$ for task $n$. 

In this work, instead of altering the architecture of the network or adding a fixed set of parameters, we seek a framework that enables all tasks to share as many parameters as feasible while retaining excellent performance, i.e., proposed task-adaptive sharing policy and gradient balance algorithm. 
We assume the layers with learnable parameters in the backbone except for the attention modules to be active layers, such as convolution and normalization layer, while other layers, such as ReLU layer and Sigmoid layer, are considered as inactive layers. 

\textbf{Layer-adaptive Sharing Policy.}
Theoretically, when manually determining whether $K$ active layers are shared or not, a combinatorial search over $2^K$ possible networks is required. 
Thus, in lieu of hand-crafted weight or layer sharing schemes, inspired by TAPS~\cite{wallingford2022task}, we relax the combinatorial issue into a learnable one and introduce a layer-adaptive sharing policy that automatically determines whether each layer of the active layers is shared or not for diverse scenes. 
Using a single weight and bias for each active layer, however, does not enable different tasks to share or monopolize the parameters dynamically at various iterations during training, hence limiting the adaptivity of OFVL-MS for the scenes.

To tackle this issue, as shown in \cref{layer-sharing}, taking a convolution layer as example, we cast the initial weight $w \in \mathbb{R}^{C_{out} \times C_{in} \times k \times k}$ 
of the convolution kernel as task-shared parameters and define two additional parameters: task-specific weight $\tilde{w} \in \mathbb{R}^{C_{out} \times C_{in} \times k \times k}$, 
and a learnable score $s \in \mathbb{R}^{1}$, where $C_{out}$, $C_{in}$ and $k$ mean output channels, input channels, and kernel size, respectively.
In forward pass, we define an indicator function for the score to judge whether the parameters of convolution layer are shared or not in current iteration, formulated as: 

\begin{equation}
	\Theta(s) = \begin{cases} 
		0 	& if \ \ s \ge \lambda \\ 
		1 	& otherwise,
	\end{cases}
	\label{indi}
\end{equation}
where $\lambda$ is a preset threshold.
The task-adaptive weight $\bar{w}$ 
used for current iteration is formulated as:

\begin{equation}
	\begin{aligned}
		\bar{w} = &\Theta(s)w + (1-\Theta(s))\tilde{w}.
	\end{aligned}
	\label{sewb}
\end{equation} 

If the score $s$ is larger than the preset threshold $\lambda$, the task-specific parameters $\tilde{w}$ will be activated and optimized, and vice versa. 
We apply above procedure on all active layers to enable different tasks to share or monopolize the parameters dynamically at various iterations. 
Besides, concluding additional parameters $\tilde{w}$ into each layer does not result in a large increase in memory cost since only the selected parameters $\bar{w}$ and $s$ are optimized at each iteration and all other parameters are kept offline.

Compared with TAPS, our proposed sharing policy delivers following merits: (1) we introduce individual task-shared weight $w$ and task-specific weight $\tilde{w}$ for each active layer rather than a coupled weight in TAPS, enabling the memory footprint to be agnostic to the number of tasks; (2) once the training for multi-task is done, the new added task can share task-shared parameters or task-specific parameters with any task in our setting, allowing for more flexible parameter sharing and real multi-task learning. 


NOtably, we set the learnable score $s$ to be task-shared so that ensuring the parameters of all scenes can be integrated into a collective model. 
Moreover, we calculate the summation of the absolute values of all scores as penalty loss to enable all tasks to share parameters as many as possible, achieving efficient storage deployment. 
Since the indicator function $\Theta(\cdot)$ is not differentiable, we need to modify its gradient during backward pass, which will be presented in Appendix 2.1. 
Notably, as illustrated in \cref{dpss}, learning task-specific batch normalization can significantly improve the localization performance while adding small parameters, so we set the parameters of normalization layers in active layers as task-specific.

\textbf{Task-specific Attention Module.}
We further embed an attention module into the backbone, empowering OFVL-MS to learn more scene-related features. 
The attention module learns a soft attention mask to the features, that can automatically determine the importance of features for each task along the channel dimension, enabling self-supervised learning of more scene-related features. 
In this work, we adopt SENet \cite{hu2018squeeze} as attention module and integrate it into the BasicBlock of each ResBlock. 
Each task $n$ has task-specific attention modules with exclusive parameters. 

\subsection{Optimize OFVL-MS}
Since each task has its own dataset domain, we need to utilize multiple GPUs to optimize these tasks.
For the sake of description, we assume that a single GPU is used to train each scene.

\textbf{Loss.}
The goal of OFVL-MS is to ensure precise visual localization for all scenes while enabling different tasks to share as many parameters as possible.
Therefore, we cast the training process of OFVL-MS as a joint optimization problem for predicted scene coordinates and scores in \cref{indi}.
For the $n$-th scene, the loss $L_{n}$ involves two terms: the scene coordinates loss $L_{n}^{sc}$ and the penalty loss $L_{n}^{pe}$. 
\begin{equation}
	L_{n} = L_{n}^{sc} + \beta L_{n}^{pe}, 
\end{equation}
where $\beta$ denotes weight coefficient used to reconcile $L_{n}^{sc}$ and $L_{n}^{pe}$.

\emph{Scene coordinates loss.} We employ the loss function proposed by KFNet \cite{zhou2020kfnet} to maximize the logarithmic likelihood for the probability density function of the predicted scene coordinates. 
Specifically, the loss function of the $n$-th scene is formatted as:
\begin{equation}
	\label{loss}
	L_{n}^{sc} = \frac{1}{Q} \sum_{i=1}^{Q} (3 log \hat{u}_{n,i} + \frac{|| d_{n,i}- \hat{d}_{n,i} ||^{2}_{2}}{2 \hat{u}_{n,i}^2} ),
\end{equation}
where $Q$ equals to $H/8 \times W/8$; $\hat{u}_{n,i}$ is the $i$-th predicted uncertainty; $d_{n,i}$ is the $i$-th ground truth scene coordinate; $\hat{d}_{n,i}$ is the $i$-th predicted scene coordinate. 

\emph{Penalty loss on the learnable scores.} 
The penalty loss $L_{n}^{pe}$ motivates all tasks to share the parameters of active layers as many as possible.
Such loss is denoted by calculating the summation of the absolute values of scores $s_{n}$ for the $n$-th scene:
\begin{equation}
	L_{n}^{pe} = \frac{1}{||S_{n}||}\sum_{s_{n}\in S_{n}}|s_{n}|,
\end{equation}
where $S_{n}$ means the collection of the scores; $||S_{n}||$ denotes the number of scores.
It is worth noting that the scores $s_{n}$ of all scenes are identical since they are set as task-shared.

\textbf{Backward Pass and Gradient Normalization Algorithm.}
For convenient portrayal, we denote the task-shared and task-specific parameters of OFVL-MS for $n$-th scene as $\chi^{sh}_{n} = \{\phi^{sh}\}$ and $\chi^{sp}_{n} = \{\phi^{sp}_{n},\theta_n\}$.

For task-specific parameters, we define the gradients of $\chi^{sp}_{n,i}$ for $n$-th scene at $i$-th iteration as: $G_{n,i}^{sp} = \nabla_{\chi^{sp}_{n,i}} L_{n,i}$, 
where $L_{n,i}$ means the loss function for the $n$-th scene at the $i$-th iteration.
Subsequently, the task-specific parameters on each GPU will be optimized based on the calculated $G_{n,i}^{sp}$.
Noting that when optimizing a scene with multiple GPUs, the gradients $G_{n,i}^{sp}$ on the GPUs would be averaged and then the parameters are updated accordingly. 
For task-shared parameters, the gradients of $\chi^{sh}_{n,i}$ for $n$-th scene at $i$-th iteration is also formulated as: $G_{n,i}^{sh} = \nabla_{\chi_{n,i}^{sh}} L_{n,i}$. 

A straightforward scheme for optimizing the task-shared parameters involves averaging the gradients $G_{n,i}^{sh}$ across all GPUs and then updating the corresponding weights. 
While this method streamlines the optimization problem, it may also trigger gradient conflict among tasks, lowering overall performance due to an unequal competition among tasks for the shared parameters, i.e., gradient magnitude disparities. 
Moreover, OFVL-MS is designed for jointly optimizing multiple indoor scenes, where the varied scene domains will further intensify the gradient conflict. 
Inspired by \cite{liu2021towards, sinha2018gradient, javaloy2021rotograd, sener2018multi}, we utilize a gradient normalization algorithm to homogenize the gradient magnitude of the task-shared parameters for all scenes, allowing all tasks to converge at same pace and alleviating the gradient conflict. 
Specifically, OFVL-MS first places gradient norms of task-shared parameters on a common scale $D$.
Considering the magnitudes and the change rate of gradient reflect whether the optimization direction in current iteration is dependable or not, we define $D$ as the linear combination of task-wise gradient magnitudes:
\begin{equation}
	D = \sum_{n=1}^{N} W_{n,i}||{G}_{n,i}^{sh}||_{2},
\end{equation}
where the weight $W_{n,i}$ is denoted as the relative convergence of each task:
\begin{equation}
	W_{n,i} = \frac{||{G}_{n,i}^{sh}||_{2} / ||{G}_{n,i-1}^{sh}||_{2}}{\sum_{j=1}^{N}||{G}_{j,i}^{sh}||_{2} / ||{G}_{j,i-1}^{sh}||_{2}}.
\end{equation}

Then, given the common scale $D$, OFVL-MS generates the optimized gradients $\hat{G}_{n,i}^{sh}$:
\begin{equation}
	\label{formule5}
	\hat{G}_{n,i}^{sh} = D\frac{{G}_{n,i}^{sh}}{||{G}_{n,i}^{sh} ||_{2}}.
 \vspace{-1.0em}
\end{equation}

Ultimately, we average the gradients $\hat{G}_{n,i}^{sh}$ on all GPUs to derive $\hat{G}_{i}^{sh}$, ensuring the gradients of the task-shared parameters for all scene are equivalent. The $\hat{G}_{i}^{sh}$ is formulated as:
\begin{equation}
	\hat{G}_{i}^{sh} = \frac{1}{N}\sum_{n=1}^{N}\hat{G}_{n,i}^{sh}.
 \vspace{-1.0em}
\end{equation}



\begin{table*} \large
	\centering
	\begin{spacing}{1.00}
		\resizebox{0.85\textwidth}{!}{%
			\begin{tabular}{l|l|ccccccc|c}
				\toprule[2pt]
				Methods & Metrics           & \multicolumn{1}{c}{Chess} & \multicolumn{1}{c}{Fire} & \multicolumn{1}{c}{Heads} & \multicolumn{1}{c}{Office} & \multicolumn{1}{c}{Pumpkin} & \multicolumn{1}{c}{Redkitchen} & \multicolumn{1}{c|}{Stairs} & \multicolumn{1}{c}{Average} \\ \midrule
                \multirow{2}{*}{\begin{tabular}[c]{@{}l@{}} AS\end{tabular} \cite{sattler2016efficient}} & \textbf{Med. Err.} & 0.03, 0.87                 & 0.02, 1.01                & 0.01, 0.82                 & 0.04, 1.15                  & 0.07, 1.69                   & 0.05, 1.72                      & 0.04, 1.01                  & 0.03, 1.18                 \\
                & \textbf{Acc.}      & ---                       & ---                    & ---                       & ---                        & ---                         & ---                            & ---                        & 68.7                         \\ \midrule
                \multirow{2}{*}{\begin{tabular}[c]{@{}l@{}} InLoc\end{tabular} \cite{taira2018inloc}} & \textbf{Med. Err.} & 0.03, 1.05                 & 0.03, 1.07                & 0.02, 1.16                 & 0.03, 1.05                  & 0.05, 1.55                   & 0.04, 1.31                      & 0.09, 2.47                  & 0.04, 1.38                 \\
                & \textbf{Acc.}      & ---                       & ---                      & ---                       & ---                        & ---                         & ---                            & ---                        & 66.3                         \\ \midrule
                \multirow{2}{*}{\begin{tabular}[c]{@{}l@{}} HLoc\end{tabular} \cite{sarlin2019coarse}} & \textbf{Med. Err.} & 0.02, 0.85                 & 0.02, 0.94                & 0.01, 0.75                 & 0.03, 0.92                  & 0.05, 1.30                   & 0.04, 1.40                      & 0.05, 1.47                  & 0.03, 1.09                 \\
                & \textbf{Acc.}      & ---                       & ---                      & ---                       & ---                        & ---                         & ---                            & ---                        & 73.1                         \\ \midrule
				\multirow{2}{*}{MS-Transformer \cite{shavit2021learning}}                                                 & \textbf{Med. Err.} & 0.11, 4.66                 & 0.24, 9.6                & 0.14, 12.19                 & 0.17, 5.66                  & 0.18, 4.44                   & 0.17, 5.94                      & 0.26, 8.45                  & 0.18, 7.27                   \\
				& \textbf{Acc.}      & ---                       & ---                      & ---                       & ---                        & ---                         & ---                            & ---                        & ---                         \\ \midrule
                \multirow{2}{*}{DSAC* \cite{brachmann2018learning}}                                                  & \textbf{Med. Err.} & 0.02, 1.10                  & 0.02, 1.24                 & 0.01, 1.82                 & 0.03, 1.15                   & 0.04, 1.34                    & 0.04, 1.68                      & 0.03, 1.16                  & \textbf{0.02}, 1.35                  \\
				& \textbf{Acc.}      & ---                      & ---                     & ---                      & ---                       & ---                        & ---                           & ---                       & 85.2                       \\ \midrule
				\multirow{2}{*}{SCoordNet \cite{zhou2020kfnet}}                                               & \textbf{Med. Err.} & 0.019, 0.63                & 0.023, 0.91               & 0.018, 1.26                & 0.026, 0.73                & 0.039, 1.09                  & 0.039, 1.18                     & 0.037, 1.06                 & 0.029, 0.98                  \\
				& \textbf{Acc.}      & ---                       & ---                      & ---                       & ---                        & ---                         & ---                            & ---                        & ---                         \\ \midrule
				\multirow{2}{*}{HSCNet \cite{li2020hierarchical}}                                                  & \textbf{Med. Err.} & 0.02, 0.7                  & 0.02, 0.9                 & 0.01, 0.9                  & 0.03, 0.8                   & 0.04, 1.0                    & 0.04, 1.2                       & 0.03, 0.8                   & 0.03, 0.9                    \\
				& \textbf{Acc.}      & \textbf{97.5}             & 96.7                     & \textbf{100.0}                       & 86.5                       & 59.9                        & 65.5                           & \textbf{87.5}                       & 84.8                        \\ \midrule
				\multirow{2}{*}{FDANet \cite{xie2022deep}}                                                  & \textbf{Med. Err.} & 0.018, 0.64                & 0.018, 0.73               & 0.013, 1.07                & 0.026, 0.75                 & 0.036, 0.91                  & 0.034, 1.03                     & 0.041, 1.14                 & 0.026, 0.89                  \\
				& \textbf{Acc.}      & 95.70                     & 96.10                    & 99.20                     & 88.08                      & 65.65                       & 78.32                          & 62.80                      & 83.69                       \\ \midrule
				\multirow{2}{*}{VS-Net \cite{huang2021vs}}                                        & \textbf{Med. Err.} & \textbf{0.015, 0.5}                & 0.019, 0.8               & 0.012, 0.7                & \textbf{0.021, 0.6}                 & 0.037, 1.0                  & 0.036, 1.1                     & 0.028, 0.8                 & 0.024, 0.8                  \\
				& \textbf{Acc.}      & ---                     & ---                    & ---                     & ---                      & ---                       & ---                          & ---                      & ---                       \\ \midrule[2pt]
				\multirow{2}{*}{OFVL-MS18}                                                & \textbf{Med. Err.} & 0.021, 0.67                & 0.018, 0.67               & 0.010, 0.56                & 0.030, 0.83                 & 0.033, 0.96                  & 0.035, 1.02                     & 0.031, 0.89                 & 0.025, 0.80                  \\
				& \textbf{Acc.}      & 96.20                     & 97.55                    & 98.90                     & 81.73                     & 67.15                       & 75.06                          & 79.80                      & 85.19                       \\ \midrule		\multirow{2}{*}{OFVL-MS34}                                                & \textbf{Med. Err.} & 0.019, 0.63                & 0.017, 0.65               & \textbf{0.008, 0.53}       & 0.027, 0.74                 & 0.031, 0.93                  & 0.032, 1.01                     & 0.027, \textbf{0.69}                 & 0.023, 0.74                  \\
				& \textbf{Acc.}      & 97.40                     & 96.60                    & \textbf{100.0}              & 85.58                      & 67.50                       & 77.14                          & 87.40             & 87.37                       \\ \midrule
				\multirow{2}{*}{OFVL-MS50}                                                & \textbf{Med. Err.} & \textbf{0.015, 0.50}       & \textbf{0.015, 0.59}      & \textbf{0.008}, 0.56                & 0.023, \textbf{0.63}        & \textbf{0.030, 0.86}         & \textbf{0.031, 0.99}            & \textbf{0.026}, 0.76                 & 0.021, \textbf{0.69}         \\
				& \textbf{Acc.}      & 97.10                     & \textbf{99.40}           & \textbf{100.0}                       & \textbf{89.53}             & \textbf{68.80}              & \textbf{81.48}                 & 84.70                       & \textbf{88.72}              \\ \bottomrule[2pt]
			\end{tabular}
		}
		\caption{The median positional error (m), rotational error ($^{\circ}$), and 5cm-5$^{\circ}$ accuracy (\%) of different methods on \textbf{7-Scenes dataset}.}
		\label{7-Scenes}
	\end{spacing}
\end{table*}

\begin{table}[]
	\centering
	\renewcommand\arraystretch{1.15}
	\resizebox{0.26\textwidth}{!}{%
		\begin{tabular}{l|cc}
			\toprule[1pt]
			\multicolumn{1}{c|}{Methods} & \multicolumn{1}{c}{Med. Err.} & \multicolumn{1}{c}{Acc.} \\ \midrule
			DSAC* \cite{brachmann2021visual}                      & ---              & 99.1                                         \\
			SCoordNet \cite{zhou2020kfnet}                   & ---              & 98.9                                        \\
			HSCNet \cite{li2020hierarchical}                      & 0.011, 0.50              & 99.3                                        \\
			FDANet \cite{xie2022deep}                      & 0.014, 0.37              & 99.6                                        \\ \midrule
			OFVL-MS18                    & 0.013, 0.48              & 98.7                                       \\
			OFVL-MS34                    & 0.007, 0.25              & 99.9                                        \\
			OFVL-MS50                    & 0.008, 0.30              & 99.5                                        \\ \bottomrule[1pt]
		\end{tabular}
	}
	\caption{The median positional error (m), rotational error ($^{\circ}$), and 5cm-5$^{\circ}$ accuracy (\%) of different methods on \textbf{12-Scenes dataset}.}
	\label{Scenes12}
\end{table}

\subsection{Pose Estimation}
We design the regression layer as a fully convolutional structure to predict dense 3D scene coordinates as well as 1D uncertainty, where the uncertainty is utilized to measure the prediction effect by quantifying the noise induced from both data and model.
Based on the predicted 2D pixel coordinates-3D scene coordinates correspondences, we apply the RANSAC-based PnP algorithm to minimize reprojection errors and finally derive camera pose $T$.

\section{Experiments}
\subsection{Datasets}
\textbf{7-Scenes} \cite{shotton2013scene} dataset records 41k RGB-D images and corresponding camera poses of seven different indoor environments using a handheld Kinect camera.
\textbf{12-Scenes} \cite{valentin2016learning} dataset, whose recorded environment is larger than that of 7-Scenes, records RGB-D images in twelve indoor environments with an iPad color camera.

\subsection{Experimental Settings}
\textbf{Implementation Details.}
We employ the Adamw solver for optimization with a weight decay of $0.05$. 
The initial learning rate is set to $1.4 \times 10^{-3}$ for 7-Scenes while $2.4 \times 10^{-3}$ for 12-Scenes with cosine annealing.
Considering the number of images for each scene is distinct, we train OFVL-MS for $200k$ iterations with batch size of $4$. 
For layer-adaptive sharing policy, we set the threshold $\lambda = 0.5$ in \cref{indi} to determine whether each active layer of the backbone is shared or not. 
Besides, we set $\beta=0.25$ in \cref{sewb} to reconcile scene coordinates loss and penalty loss.
More implementation details can be found in Appendix 2.

\textbf{Evaluation Metrics.} 
Following previous works~\cite{xie2022deep, zhou2020kfnet, li2020hierarchical}, we evaluate our method using the following metrics: 
(i) the median positional and rotational errors of the predicted pose; 
(ii) the percentage of images with positional and rotational errors less than 5cm and 5$^{\circ}$. 
\subsection{Comparison with State-of-the-art Methods}
We design three versions OFVL-MS18, OFVL-MS34 and OFVL-MS50 of our method by using ResNet18, ResNet34, ResNet50 \cite{he2016deep} as backbone respectively, and then compare OFVL-MS families with other state-of-the-arts on 7-Scenes and 12-Scenes datasets, with the results reported in \cref{7-Scenes} and \cref{Scenes12}.

\begin{figure}
	\begin{minipage}{0.495\linewidth}
		\includegraphics[width=1.0\hsize]{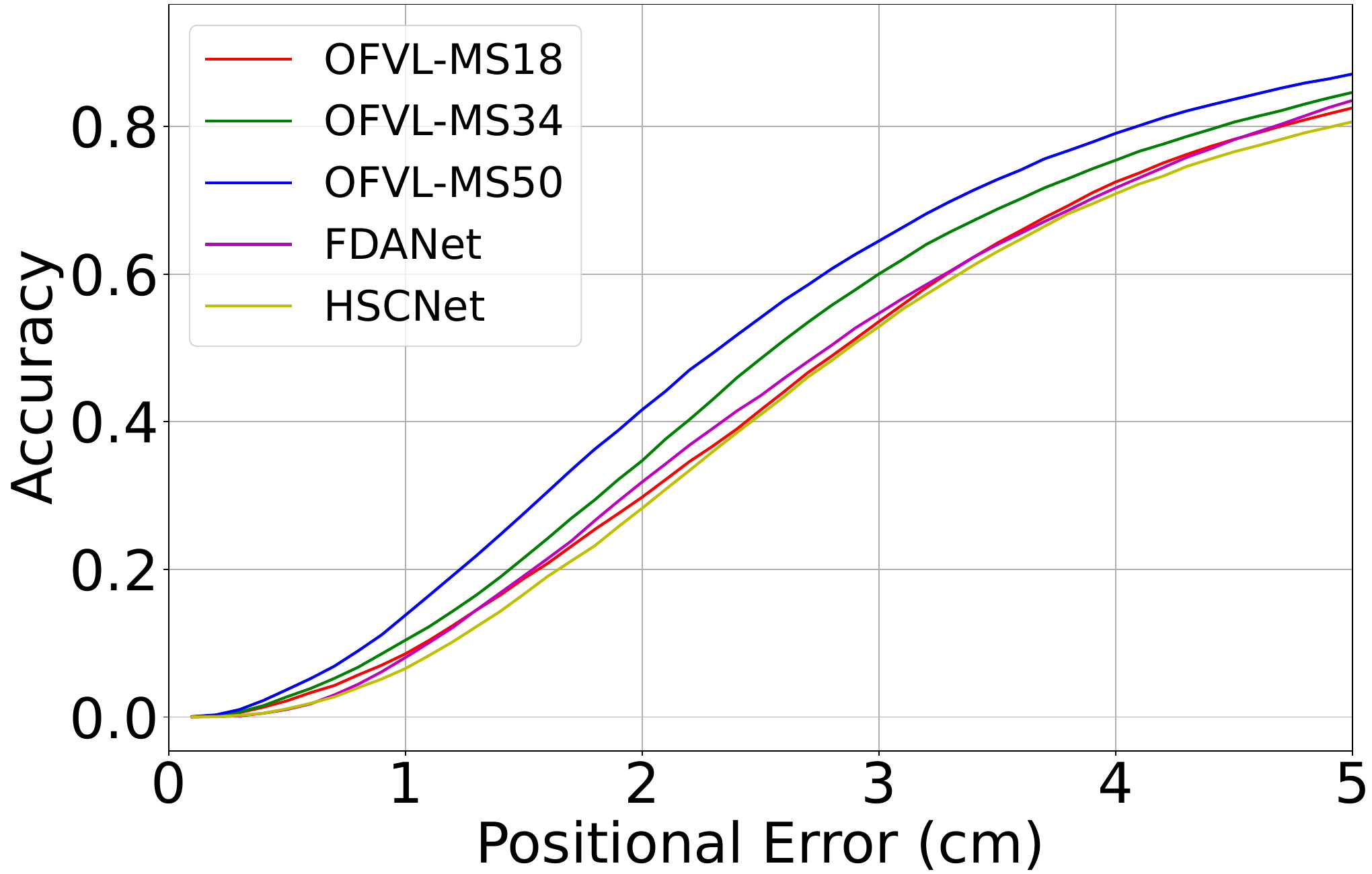}
	\end{minipage}
	\hfill
	\begin{minipage}{0.495\linewidth}
		\includegraphics[width=1.0\hsize]{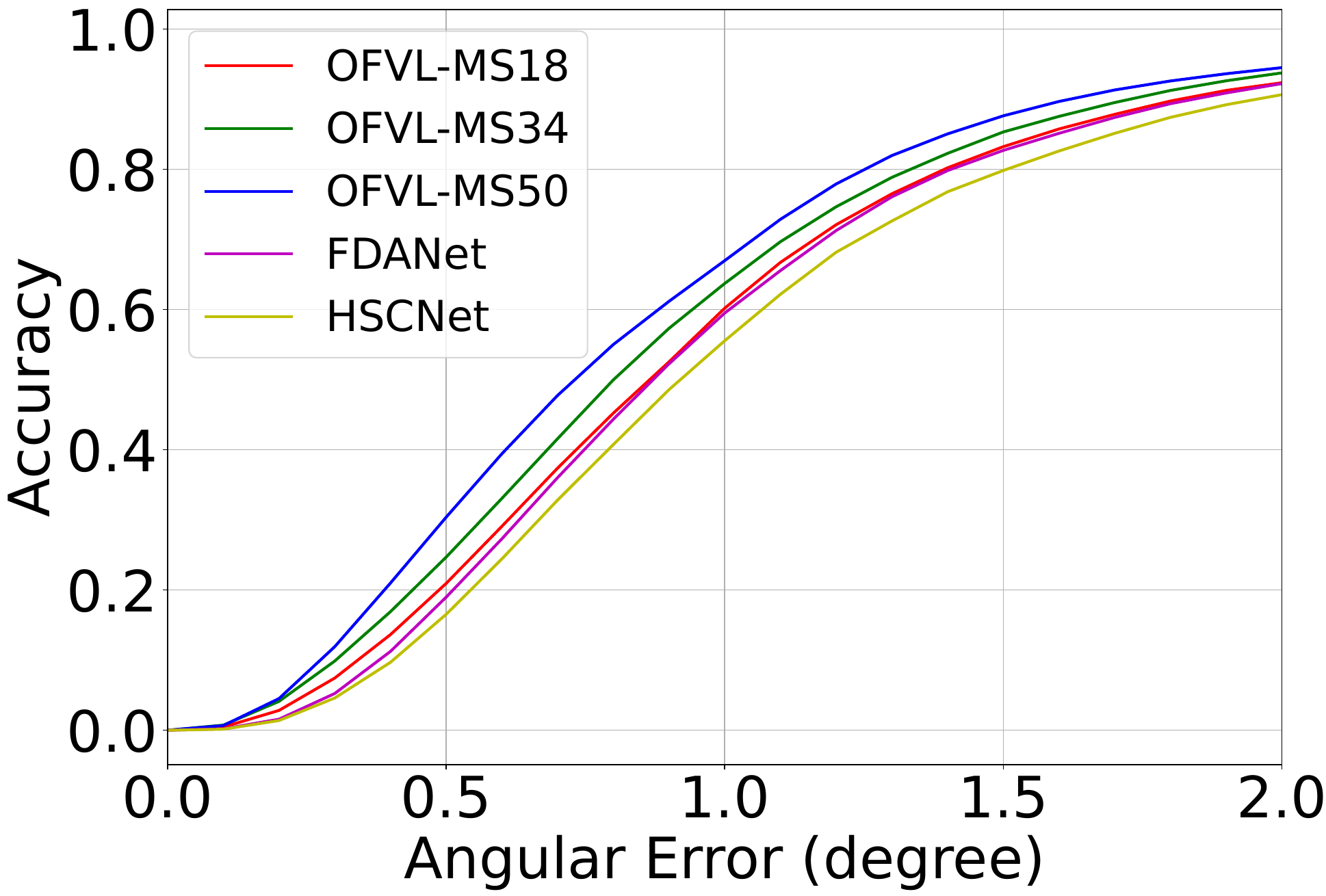}
	\end{minipage}
	\caption{\textbf{The cumulative pose errors distribution} of different methods on 7-Scenes dataset. We aggregate the poses across all scenes and calculate the percentage of poses with the error threshold increasing.} 
	\label{7Scenes_AUC}
\end{figure}

\textbf{Localization on 7-Scenes. }
We compare OFVL-MS with representative structure-based methods (AS \cite{sattler2016efficient}, InLoc \cite{taira2018inloc}, HLoc \cite{sarlin2019coarse}), APR methods (MS-Transformer \cite{shavit2021learning}), and SCoRe-based methods (DSAC* \cite{brachmann2021visual}, SCoordNet \cite{zhou2020kfnet}, HSCNet \cite{li2020hierarchical}, FDANet \cite{xie2022deep}, and VSNet \cite{huang2021vs}).
As shown in \cref{7-Scenes}, OFVL-MS surpasses existing methods by non-trivial margins in terms of all evaluation metrics.
Specifically, OFVL-MS18/34 outperforms the structure-based method HLoc by $12.09\%/14.27\%$ in terms of 5cm-5$^{\circ}$ accuracy.
Besides, compared with SCoRe-based methods HSCNet and FDANet, OFVL-MS18/34 realizes outstanding performance with the improvement of $0.4\%/2.57\%$ and $1.51\%/3.68\%$.
Compared with the cutting-edge method VS-Net, OFVL-MS18/34 also achieve higher performance. 
Moreover, OFVL-MS50 yields $0.021$m median position error, $0.69^{\circ}$ median rotational error and $88.72\%$ 5cm-5$^{\circ}$ accuracy, establishing a new state-of-the-art for 7-Scenes dataset. 
\cref{7Scenes_AUC} shows the cumulative pose errors distribution of different approaches on 7-Scenes dataset, which further demonstrates the superiority of OFVL-MS families in visual localization. 

\textbf{Localization on 12-Scenes. }
As illustrated in \cref{Scenes12}, we compare OFVL-MS families with state-of-the-arts on 12-Scenes dataset.
It can be observed that all methods achieve excellent results since the training trajectories closely resemble the test trajectories. 
Despite this, OFVL-MS families exhibit exceptional performances, in which OFVL-MS34 realizes the most superior performance with the positional errors of $7$mm and localization accuracy of $99.9\%$.

\begin{table}[]
	\centering
	\resizebox{0.35\textwidth}{!}{%
		\begin{tabular}{lcccc}
			\toprule[2pt]
			\multicolumn{1}{l|}{Methods}  & Total Params (M) & Med. Err. & Acc. \\ \midrule
			\multicolumn{4}{c}{7-Scenes} \\ \midrule
			\multicolumn{1}{l|}{DSAC++ \cite{brachmann2018learning}} & 182.384 & 0.036, 1.10 & 74.4 \\
			\multicolumn{1}{l|}{HSCNet \cite{li2020hierarchical}} & 288.751 & 0.030, 0.90 & 84.8 \\
			\multicolumn{1}{l|}{FDANet \cite{xie2022deep}} & 168.758 & 0.026, 0.89 & 83.69 \\ \midrule
			
			\multicolumn{1}{l|}{OFVL-MS18}  & \textbf{48.803}    & 0.025, 0.80          & 85.19     \\
			\multicolumn{1}{l|}{OFVL-MS34}  & 64.403             & 0.023, 0.74          & 87.37     \\
			\multicolumn{1}{l|}{OFVL-MS50}  & 53.015             & \textbf{0.021, 0.69}          & \textbf{88.72}     \\ \midrule
			\multicolumn{4}{c}{12-Scenes} \\ \midrule
			\multicolumn{1}{l|}{HSCNet \cite{li2020hierarchical}} & 495.002 & 0.011, 0.50 & 99.3 \\
			\multicolumn{1}{l|}{FDANet \cite{xie2022deep}} & 289.299 & 0.014, 0.37 & 99.6 \\ \midrule
			
			\multicolumn{1}{l|}{OFVL-MS18}  & \textbf{75.453}   & 0.013, 0.48         & 98.7     \\
			\multicolumn{1}{l|}{OFVL-MS34}  & 158.693            & \textbf{0.007, 0.25}          & \textbf{99.9}     \\
			\multicolumn{1}{l|}{OFVL-MS50}  & 126.694            & 0.008, 0.30          & 99.5    \\ \bottomrule[2pt]
		\end{tabular}
	}
	\caption{\textbf{The model size of different methods.} OFVL-MS families achieve the best localization accuracy with much less parameters.}
	\label{storage}
\end{table}

\textbf{Model Size Comparison.}
We compare the storage space occupied by different methods to demonstrate the efficient storage deployment of OFVL-MS families. 
Previous works typically train a separate model for each scene, resulting in a linear increase in model size with the number of scenes. 
However, OFVL-MS deposits multiple models with a majority of shared parameters into a single one, realizing efficient storage.
As shown in \cref{storage}, OFVL-MS families reduce the model parameters significantly compared with other state-of-the-arts. 
For 7-Scenes dataset, the parameters size of OFVL-MS50 is only $1/5$ of that of HSCNet, but the localization accuracy is improved by $3.92\%$.
For 12-Scenes dataset, OFVL-MS34 achieves the best performance with much fewer parameters (only $1/3$ of HSCNet).

\subsection{Joint Training vs Separate Training} \label{joint_separate}
To further demonstrate the efficiency of jointly optimizing localization tasks across scenes, we train a separate model for each scene. 
We choose OFVL-MS34 as the benchmark for validation. 
As shown in \cref{Separate}, OFVL-MS34 reduces total model size from $177.779$M to $64.403$M by sharing parameters for all scenes. 
Besides, it is astonishing to find OFVL-MS34 achieves competitive performance through joint training, indicating that closely-related tasks have mutual benefits.
\begin{table}[]
	\centering
	\resizebox{0.46\textwidth}{!}{%
		\begin{tabular}{l|cccc}
			\toprule[1.2pt]
			Methods      & Total Params (M) & Med. Err. & Acc. \\ \midrule
			Separate Learning   & 177.779                             & \textbf{0.023, 0.74}          & 86.50     \\
			Joint Learning & \textbf{64.403}                              & \textbf{0.023, 0.74}         & \textbf{87.37}      \\   \bottomrule[1.2pt]
		\end{tabular}
	}
	\caption{The comparison between \textbf{joint and separate training of OFVL-MS34} on 7-Scenes dataset.}
	\label{Separate}
\end{table}

\begin{table}[]
	\centering
	\resizebox{0.43\textwidth}{!}{%
		\begin{tabular}{l|cccc}
			\toprule[1.5pt]
			Methods      & Total Params (M) & Med. Err. & Acc. \\ \midrule
			EXP1   & \textbf{50.243}                             & 0.027, 0.82          & 79.94     \\
			EXP2 & 64.391                              & 0.024, 0.76         & 86.10      \\ 
			EXP3 (Ours)   & 64.403                          & \textbf{0.023, 0.74}           & \textbf{87.37}     \\\bottomrule[1.5pt]
		\end{tabular}
	}
	\caption{The comparison between \textbf{different parameters sharing strategies} on 7-Scenes dataset.}
	\label{diverse_sharing}
\end{table}
\subsection{Diverse Parameters Sharing Strategies} \label{dpss}
To verify the effectiveness of the proposed layer-adaptive sharing policy, we apply three different parameter sharing strategies on OFVL-MS34 for 7-Scenes dataset. 
EXP1: All parameters of active layers are set as task-shared.
EXP2: All parameters of active layers (both convolutional and batch normalization layers) are determined whether to be shared by scores.
EXP3: All parameters of active layers (only convolutional layers) are determined whether to be shared by scores, and the batch normalization layers are set as task-specific.
As shown in \cref{diverse_sharing}, compared to setting all parameters as task-shared, OFVL-MS34 significantly improves localization performance from $79.94$ to $87.37$ in terms of 5cm-5$^{\circ}$ accuracy at the expense of a small increase in model parameters, indicating that using additional task-specific parameters to learn scene-related features is critical to resolve gradient conflict. 
Besides, the performance of OFVL-MS is further enhanced with BN layers set as task-specific.

\begin{table}
	\centering
	\resizebox{0.35\textwidth}{!}{%
		\begin{tabular}{l|ccc}
			\toprule[1.6pt]
			Methods & Increased Params (M)   & Med. Err.            & Acc.              \\ \midrule
			\multicolumn{4}{c}{EXP1: 12-Scenes to 7-Scenes} \\ \midrule
			HSCNet \cite{li2020hierarchical}    & 41.250 $\times$ 7  & 0.030, 0.90             & 84.80	                \\
			FDANet \cite{xie2022deep}    & 24.108 $\times$ 7  & 0.026, 0.89           & 83.69               \\
			OFVL-MS18$^{\dagger}$ & \textbf{5.476 $\times$ 7}  & 0.029, 0.93           & 77.59                 \\
			OFVL-MS34$^{\dagger}$ & 12.117 $\times$ 7  & 0.023, 0.75           & 85.73                 \\
			OFVL-MS50$^{\dagger}$ & 9.881 $\times$ 7  & \textbf{0.021, 0.71}  & \textbf{86.75} \\ \midrule
			\multicolumn{4}{c}{EXP2: 7-Scenes to 12-Scenes} \\ \midrule
			HSCNet \cite{li2020hierarchical}    & 41.250 $\times$ 12 & 0.011, 0.50             & 99.3	 \\ 
			FDANet \cite{xie2022deep}    & 24.108 $\times$ 12 & 0.014, 0.37           & \textbf{99.6}               \\ 
			OFVL-MS18$^{\dagger}$ & \textbf{5.597 $\times$ 12} & 0.009, 0.38          & 96.7                 \\
			OFVL-MS34$^{\dagger}$ & 6.501 $\times$ 12 & 0.009, 0.31          & 97.3                 \\
			OFVL-MS50$^{\dagger}$ & 5.835 $\times$ 12 & \textbf{0.008, 0.29}  		  & 98.3  \\\bottomrule[1.6pt]
		\end{tabular}
	}
	\renewcommand\arraystretch{1.5}
	\caption{\textbf{Experiments of generalizing to new scenes}. ${\dagger}$ indicates using the models trained on 12-Scenes/7-Scenes to conduct the generalization experiments on 7-Scenes/12-Scenes.
	Increased Params (M) means the extra parameters size that each method requires when generalizing to new scenes.}
	\label{incremental}
\end{table}

\subsection{Generalize to New Scenes} \label{incremen}
In this part, we conduct two experiments to demonstrate that OFVL-MS can generalize to new scenes with much fewer parameters and thus can scale up gracefully with the number of scenes. 
We utilize the model trained on 12-Scenes/7-Scenes and conduct the generalization experiments on 7-Scenes/12-Scenes. 
Specifically, we freeze the task-shared parameters trained on 12-Scenes/7-Scenes, and add task-specific parameters as well as an additional regression layer for each scene of 7-Scenes/12-Scenes to predict the scene coordinates.

As shown in \cref{incremental}, despite generalizing to a new scene, OFVL-MS34/50 still outperform HSCNet and FDANet by $0.93\%/1.95\%$ and $2.04\%/3.06\%$ in terms of 5cm-5$^{\circ}$ accuracy for EXP1, illustrating that OFVL-MS can avoid catastrophic forgetting and achieve genuine incremental learning.
Besides, compared with $41.250/24.108$ M increased parameters of HSCNet and FDANet, OFVL-MS18/34/50 only need $5.476/12.117/9.881$ M parameters when generalizing to a new scene, realizing efficient storage.
 
For EXP2, OFVL-MS families yield the lowest localization errors.
It is worth noting that the incremental models achieve more precise localization performance in most of scenes except for Floor5b, resulting in the 5cm-5$^{\circ}$ accuracy declined, which will be presented in Appendix 3.
Moreover, OFVL-MS families realize efficient storage deployment with $5.597/6.501/5.835$ M additional parameters compared with HSCNet and FDANet. 

\subsection{Ablation study} \label{section4_7}
To comprehensively confirm the veracity of the modules suggested in this work, various variants of OFVL-MS34 are validated using the 7-Scenes dataset. 
As shown in \cref{modules}, all of the components contribute to outstanding performance.
EXP1: Removing all task-specific attention modules results in a large drop in localization accuracy, demonstrating the strong ability of TSAM to generate more scene-related features, realizing efficient scene parsing. 
EXP2: Removing gradient normalization algorithm leads to much lower accuracy, validating that homogenizing the gradient magnitude of the task-shared parameters alleviates the gradient conflict significantly.
EXP3: Removing penalty loss results in degraded localization accuracy, indicating that promoting the informative parameters sharing across scenes improves localization performance.
\begin{table}[]
	\centering
	\resizebox{0.46\textwidth}{!}{%
		\begin{tabular}{cccccccc}
			\toprule[1pt]
			Methods  & TSAM & GNA  & PL & Med. Err.  & Acc. & Params-t (M)           \\ \midrule
			EXP1           &\xmark     &\cmark      &\cmark  & 0.026, 0.79                 & 84.30 & \textbf{63.297}               \\ 
			EXP2          &\cmark     &\xmark      &\cmark  & 0.025, 0.77                 & 84.16 & 64.403              \\
			EXP3          &\cmark     &\cmark      &\xmark  & \textbf{0.023, 0.74}                 & 86.25  & 78.059\\
			EXP4           &\cmark     &\cmark      &\cmark  & \textbf{0.023, 0.74} & \textbf{87.37} & 64.403\\             \bottomrule[1pt]
		\end{tabular}%
	}
	\caption{\textbf{Ablation study with various variants of OFVL-MS} on 7-Scenes dataset. TSAM: Task-specific Attention Module, GNA: Gradient Normalization Algorithm, PL: Penalty Loss. Params-t means the total parameters of OFVL-MS34 for the seven scenes.}
	\label{modules}
\end{table}

\begin{figure}
	\centering
	\includegraphics[width=0.83\hsize]{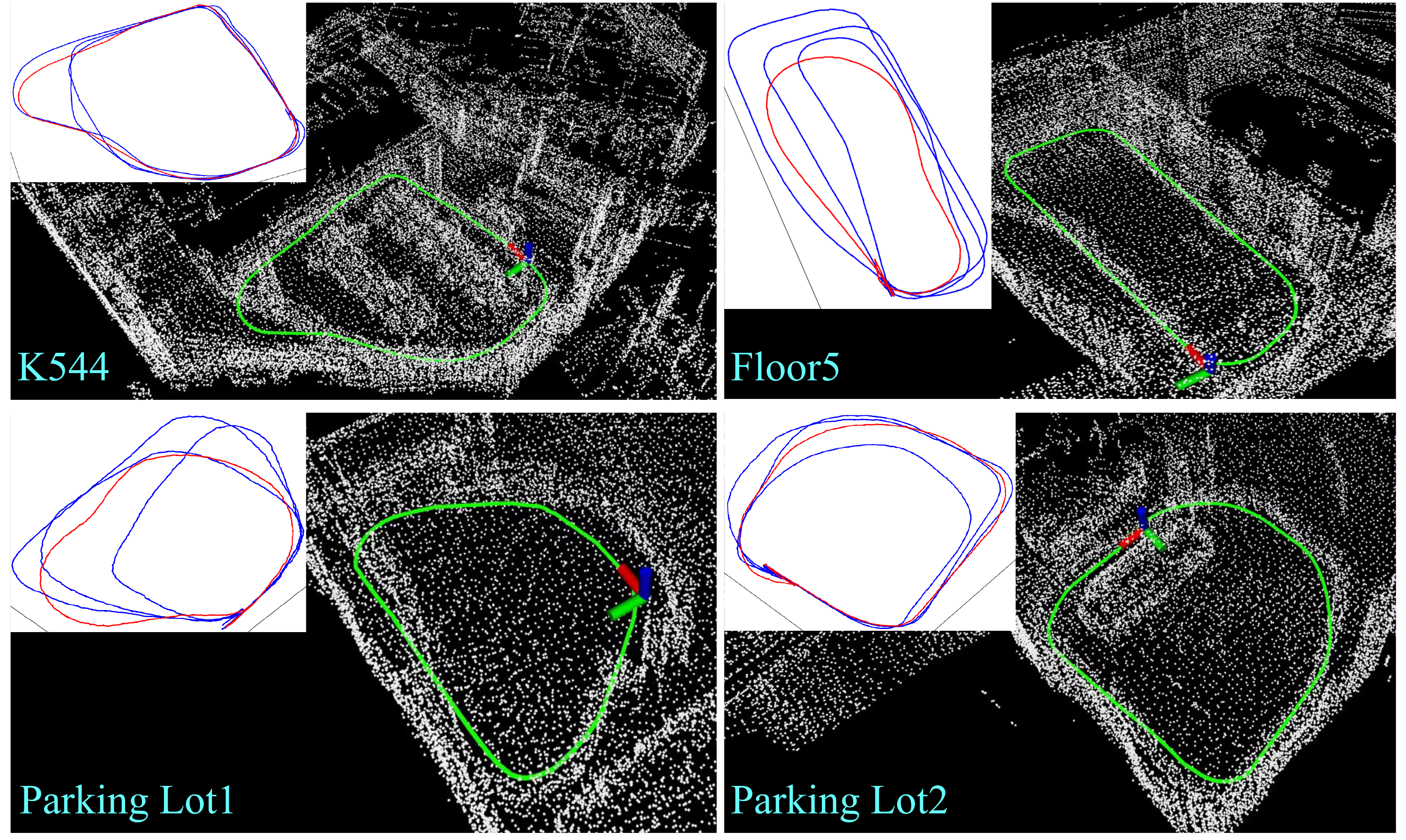}
	\caption{\textbf{LIVI dataset.} The blue lines indicate training trajectories whereas
		the red lines indicate test trajectories.}
	\label{real_scene}
\end{figure}

\begin{table}[] \Huge
	\centering
	\renewcommand\arraystretch{1.5}
	\resizebox{0.42\textwidth}{!}{%
		\begin{tabular}{l|c|c|c|c|c|c|c}
			\toprule[3pt]
			Methods                    &  Metric         & \multicolumn{1}{c|}{K544} & \multicolumn{1}{c|}{Floor5} & \multicolumn{1}{c|}{Parking lot1} & \multicolumn{1}{c|}{Parking lot2} & Average & Params-t (M)\\ \midrule
            \multirow{2}{*}{SCoordNet~\cite{zhou2020kfnet}} & Med. Err. & 0.171, 2.12     & 0.208, 1.94        & 0.353, 2.97      & 0.184, 2.13      & 0.229, 2.29 &
            \multirow{2}{*}{93.086}\\  & Acc.  & 9.86 & 20.31 & 11.28 & 20.82 & 15.57 \\ \midrule
            \multirow{2}{*}{FDANet~\cite{xie2022deep}} & Med. Err. & 0.143, 1.89     & 0.167, 1.59        & 0.291, 2.75      & 0.138, 1.61      & 0.185, 1.96  & \multirow{2}{*}{96.432}\\  & Acc.  & 12.64 & 23.87 & 13.31 & 22.89 & 18.18 \\ \midrule
			\multirow{2}{*}{OFVL-MS18} & Med. Err. & 0.074, 1.12     & 0.174, 1.84        & 0.274, 2.28      & 0.100, 1.31      & 0.155, 1.64  & \multirow{2}{*}{\textbf{37.246}} \\ & Acc.  & 39.21 & 16.92 & 24.50 & 28.34 & 27.24 \\ \midrule
			\multirow{2}{*}{OFVL-MS34} & Med. Err. & 0.071, 1.01                         & \textbf{0.147}, 1.48                           & 0.278, \textbf{2.08}                          & \textbf{0.095}, 1.17      & 0.147, 1.43 & \multirow{2}{*}{73.184} \\ & Acc.  & 42.53 & 25.54 & 25.72 & \textbf{29.03} & 30.71 \\ \midrule
			\multirow{2}{*}{OFVL-MS50} & Med. Err. & \textbf{0.050, 0.81}                         & 0.148, \textbf{1.37}                           & \textbf{0.265}, 2.48                          & 0.107, \textbf{1.02}                         & \textbf{0.142, 1.42} & \multirow{2}{*}{45.678} \\ & Acc.  & \textbf{49.91} & \textbf{30.72} & \textbf{26.17} & 28.34 & \textbf{33.79} \\ \bottomrule[3pt]
		\end{tabular}
	}
	\caption{The median positional error (m) and rotational error ($^{\circ}$) of OFVL-MS families on \textbf{LIVL dataset}. 
    Params-t means that the total parameters of OFVL-MS for the four scenes. }
	\label{realsense_table}
\end{table}

\subsection{Camera Localization on LIVI}
Despite the existence of publicly available datasets for visual localization, there is no dataset for large-scale indoor scenes. 
Thus, we introduce the challenging \textbf{LIVL} dataset containing RGB-D images tagged with 6-DoF camera poses collected around four scenes. 
(i) \textbf{K544}: spanning about $12 \times 9$m$^{2}$.
(ii) \textbf{Floor5}: spanning about $12 \times 5$m$^{2}$.
(iii) \textbf{Parking lot1} spanning about $8 \times 6$m$^{2}$.
(iv) \textbf{Parking lot2} spanning about $8 \times 8$m$^{2}$. 
Each scene contains three sequences for training and one sequence for test. 
A massive proportion of motion blur and sparse texture in the scene make visual localization in the four scenes challenging. 
We give the visualization of \textbf{LIVL} dataset in \cref{real_scene}.
The dataset was collected using a autonomous platform armed with a RealSense D435 camera and a VLP-16 laser radar.
The RGB and depth images are captured at a resolution of $640 \times 480$ pixels and aligned with point clouds using timestamp.
We utilize the LiDAR-based SLAM system A-LOAM \cite{zhang2015visual} to compute the ground truth pose.
More details of the dataset can be found in Appendix 4.

As shown in \cref{realsense_table}, we can observe that OFVL-MS50 realizes the best performance with $0.142$m and $1.42^{\circ}$ median localization error.
Wherein, OFVL-MS50 yields $0.05$m and $0.81^{\circ}$ localization error in K544 scene that contains discriminative texture. 
Moreover, Floor5 and Parking lot1 are laborious for OFVL-MS families to localize since there exists repetitive and sparse texture, and illumination disturbance. 
Besides, we can also observe that 5cm-5$^{\circ}$ accuracy is inferior due to the large scale of LIVI dataset. 
Compared with typical SCoRe based methods SCoordNet~\cite{zhou2020kfnet} and FDANet~\cite{xie2022deep}, OFVL-MS families outperform them by non-trivial margins in terms of all evaluation metrics while necessitating much fewer total parameters, further indicating that the closely-related tasks benefit from the shared parameters and the efficacy of our OFVL-MS. 


\section{Conclusion}
In this work, we introduce OFVL-MS, a unified network that achieves precise visual localization across scenes in a multi-task learning manner.
OFVL-MS achieves high performance for all tasks and keeps storage efficient for model deployment through forward pass (layer-adaptive sharing policy) and backward pass (gradient normalization algorithm) of the network. 
Moreover, a penalty loss is proposed to motivate OFVL-MS to share parameters as many as possible while maintaining precise localization accuracy. 
We demonstrate that OFVL-MS can generalize to a new scene with small task-specific parameters while realizing superior localization performance. 
We also publish a \textbf{new large indoor dataset LIVL} to provide a new test benchmark for the community. 

\textbf{Acknowledgement}.
This work was supported in part by National Natural Science Foundation of China under Grant 62073101,
in part by Science and Technology Innovation Venture Capital Project of Tiandi Technology Co., LTD. (2022-2-TD-QN009), 
in part by “Ten Thousand Million” Engineering Technology Major Special Support Action Plano of Heilongjiang Province, China (SC2021ZX02A0040), and in part by Self-Planned Task (SKLRS202301A09) of SKLRS (HIT) of China.

{\small
	\bibliographystyle{ieee_fullname}
	\bibliography{egbib}

\begin{thebibliography}{10}\itemsep=-1pt

\bibitem{abouelnaga2021distillpose}
Yehya Abouelnaga, Mai Bui, and Slobodan Ilic.
\newblock Distillpose: Lightweight camera localization using auxiliary
  learning.
\newblock In {\em 2021 IEEE/RSJ International Conference on Intelligent Robots
  and Systems (IROS)}, pages 7919--7924. IEEE, 2021.

\bibitem{arandjelovic2016netvlad}
Relja Arandjelovic, Petr Gronat, Akihiko Torii, Tomas Pajdla, and Josef Sivic.
\newblock Netvlad: Cnn architecture for weakly supervised place recognition.
\newblock In {\em Proceedings of the IEEE conference on computer vision and
  pattern recognition}, pages 5297--5307, 2016.

\bibitem{brachmann2018learning}
Eric Brachmann and Carsten Rother.
\newblock Learning less is more-6d camera localization via 3d surface
  regression.
\newblock In {\em Proceedings of the IEEE conference on computer vision and
  pattern recognition}, pages 4654--4662, 2018.

\bibitem{brachmann2021visual}
Eric Brachmann and Carsten Rother.
\newblock Visual camera re-localization from rgb and rgb-d images using dsac.
\newblock {\em IEEE transactions on pattern analysis and machine intelligence},
  44(9):5847--5865, 2021.

\bibitem{campos2021orb}
Carlos Campos, Richard Elvira, Juan J~G{\'o}mez Rodr{\'\i}guez, Jos{\'e}~MM
  Montiel, and Juan~D Tard{\'o}s.
\newblock Orb-slam3: An accurate open-source library for visual,
  visual--inertial, and multimap slam.
\newblock {\em IEEE Transactions on Robotics}, 37(6):1874--1890, 2021.

\bibitem{chen2018gradnorm}
Zhao Chen, Vijay Badrinarayanan, Chen-Yu Lee, and Andrew Rabinovich.
\newblock Gradnorm: Gradient normalization for adaptive loss balancing in deep
  multitask networks.
\newblock In {\em International conference on machine learning}, pages
  794--803. PMLR, 2018.

\bibitem{chen2020just}
Zhao Chen, Jiquan Ngiam, Yanping Huang, Thang Luong, Henrik Kretzschmar, Yuning
  Chai, and Dragomir Anguelov.
\newblock Just pick a sign: Optimizing deep multitask models with gradient sign
  dropout.
\newblock {\em Advances in Neural Information Processing Systems},
  33:2039--2050, 2020.

\bibitem{chennupati2019multinet++}
Sumanth Chennupati, Ganesh Sistu, Senthil Yogamani, and Samir A~Rawashdeh.
\newblock Multinet++: Multi-stream feature aggregation and geometric loss
  strategy for multi-task learning.
\newblock In {\em Proceedings of the IEEE/CVF Conference on Computer Vision and
  Pattern Recognition Workshops}, pages 0--0, 2019.

\bibitem{dai2023oamatcher}
Kun Dai, Tao Xie, Ke Wang, Zhiqiang Jiang, Ruifeng Li, and Lijun Zhao.
\newblock Oamatcher: An overlapping areas-based network for accurate local
  feature matching.
\newblock {\em arXiv preprint arXiv:2302.05846}, 2023.

\bibitem{dai2023eaainet}
Kun Dai, Tao Xie, Ke Wang, Zhiqiang Jiang, Dedong Liu, Ruifeng Li, and Jiahe
  Wang.
\newblock Eaainet: An element-wise attention network with global affinity
  information for accurate indoor visual localization.
\newblock {\em IEEE Robotics and Automation Letters}, 8(6):3166--3173, 2023.

\bibitem{ding2019camnet}
Mingyu Ding, Zhe Wang, Jiankai Sun, Jianping Shi, and Ping Luo.
\newblock Camnet: Coarse-to-fine retrieval for camera re-localization.
\newblock In {\em Proceedings of the IEEE/CVF International Conference on
  Computer Vision}, pages 2871--2880, 2019.

\bibitem{doersch2017multi}
Carl Doersch and Andrew Zisserman.
\newblock Multi-task self-supervised visual learning.
\newblock In {\em Proceedings of the IEEE international conference on computer
  vision}, pages 2051--2060, 2017.

\bibitem{dosovitskiy2020image}
Alexey Dosovitskiy, Lucas Beyer, Alexander Kolesnikov, Dirk Weissenborn,
  Xiaohua Zhai, Thomas Unterthiner, Mostafa Dehghani, Matthias Minderer, Georg
  Heigold, Sylvain Gelly, et~al.
\newblock An image is worth 16x16 words: Transformers for image recognition at
  scale.
\newblock {\em arXiv preprint arXiv:2010.11929}, 2020.

\bibitem{fifty2021efficiently}
Chris Fifty, Ehsan Amid, Zhe Zhao, Tianhe Yu, Rohan Anil, and Chelsea Finn.
\newblock Efficiently identifying task groupings for multi-task learning.
\newblock {\em Advances in Neural Information Processing Systems},
  34:27503--27516, 2021.

\bibitem{fischler1981random}
Martin~A Fischler and Robert~C Bolles.
\newblock Random sample consensus: a paradigm for model fitting with
  applications to image analysis and automated cartography.
\newblock {\em Communications of the ACM}, 24(6):381--395, 1981.

\bibitem{guan2021scene}
Peiyu Guan, Zhiqiang Cao, Junzhi Yu, Chao Zhou, and Min Tan.
\newblock Scene coordinate regression network with global context-guided
  spatial feature transformation for visual relocalization.
\newblock {\em IEEE Robotics and Automation Letters}, 6(3):5737--5744, 2021.

\bibitem{guo2018dynamic}
Michelle Guo, Albert Haque, De-An Huang, Serena Yeung, and Li Fei-Fei.
\newblock Dynamic task prioritization for multitask learning.
\newblock In {\em Proceedings of the European conference on computer vision
  (ECCV)}, pages 270--287, 2018.

\bibitem{he2016deep}
Kaiming He, Xiangyu Zhang, Shaoqing Ren, and Jian Sun.
\newblock Deep residual learning for image recognition.
\newblock In {\em Proceedings of the IEEE conference on computer vision and
  pattern recognition}, pages 770--778, 2016.

\bibitem{hu2018squeeze}
Jie Hu, Li Shen, and Gang Sun.
\newblock Squeeze-and-excitation networks.
\newblock In {\em Proceedings of the IEEE conference on computer vision and
  pattern recognition}, pages 7132--7141, 2018.

\bibitem{huang2021vs}
Zhaoyang Huang, Han Zhou, Yijin Li, Bangbang Yang, Yan Xu, Xiaowei Zhou, Hujun
  Bao, Guofeng Zhang, and Hongsheng Li.
\newblock Vs-net: Voting with segmentation for visual localization.
\newblock In {\em Proceedings of the IEEE/CVF Conference on Computer Vision and
  Pattern Recognition}, pages 6101--6111, 2021.

\bibitem{javaloy2021rotograd}
Adri{\'a}n Javaloy and Isabel Valera.
\newblock Rotograd: Gradient homogenization in multitask learning.
\newblock {\em arXiv preprint arXiv:2103.02631}, 2021.

\bibitem{kendall2016modelling}
Alex Kendall and Roberto Cipolla.
\newblock Modelling uncertainty in deep learning for camera relocalization.
\newblock In {\em 2016 IEEE international conference on Robotics and Automation
  (ICRA)}, pages 4762--4769. IEEE, 2016.

\bibitem{kendall2018multi}
Alex Kendall, Yarin Gal, and Roberto Cipolla.
\newblock Multi-task learning using uncertainty to weigh losses for scene
  geometry and semantics.
\newblock In {\em Proceedings of the IEEE conference on computer vision and
  pattern recognition}, pages 7482--7491, 2018.

\bibitem{kendall2015posenet}
Alex Kendall, Matthew Grimes, and Roberto Cipolla.
\newblock Posenet: A convolutional network for real-time 6-dof camera
  relocalization.
\newblock In {\em Proceedings of the IEEE international conference on computer
  vision}, pages 2938--2946, 2015.

\bibitem{lepetit2009epnp}
Vincent Lepetit, Francesc Moreno-Noguer, and Pascal Fua.
\newblock Epnp: An accurate o (n) solution to the pnp problem.
\newblock {\em International journal of computer vision}, 81(2):155--166, 2009.

\bibitem{li2020hierarchical}
Xiaotian Li, Shuzhe Wang, Yi Zhao, Jakob Verbeek, and Juho Kannala.
\newblock Hierarchical scene coordinate classification and regression for
  visual localization.
\newblock In {\em Proceedings of the IEEE/CVF Conference on Computer Vision and
  Pattern Recognition}, pages 11983--11992, 2020.

\bibitem{liu2021conflict}
Bo Liu, Xingchao Liu, Xiaojie Jin, Peter Stone, and Qiang Liu.
\newblock Conflict-averse gradient descent for multi-task learning.
\newblock {\em Advances in Neural Information Processing Systems},
  34:18878--18890, 2021.

\bibitem{liu2021towards}
Liyang Liu, Yi Li, Zhanghui Kuang, J Xue, Yimin Chen, Wenming Yang, Qingmin
  Liao, and Wayne Zhang.
\newblock Towards impartial multi-task learning.
\newblock ICLR, 2021.

\bibitem{liu2019end}
Shikun Liu, Edward Johns, and Andrew~J Davison.
\newblock End-to-end multi-task learning with attention.
\newblock In {\em Proceedings of the IEEE/CVF conference on computer vision and
  pattern recognition}, pages 1871--1880, 2019.

\bibitem{maninis2019attentive}
Kevis-Kokitsi Maninis, Ilija Radosavovic, and Iasonas Kokkinos.
\newblock Attentive single-tasking of multiple tasks.
\newblock In {\em Proceedings of the IEEE/CVF Conference on Computer Vision and
  Pattern Recognition}, pages 1851--1860, 2019.

\bibitem{moran2021deep}
Dror Moran, Hodaya Koslowsky, Yoni Kasten, Haggai Maron, Meirav Galun, and
  Ronen Basri.
\newblock Deep permutation equivariant structure from motion.
\newblock In {\em Proceedings of the IEEE/CVF International Conference on
  Computer Vision}, pages 5976--5986, 2021.

\bibitem{mur2017orb}
Raul Mur-Artal and Juan~D Tard{\'o}s.
\newblock Orb-slam2: An open-source slam system for monocular, stereo, and
  rgb-d cameras.
\newblock {\em IEEE transactions on robotics}, 33(5):1255--1262, 2017.

\bibitem{radwan2018vlocnet++}
Noha Radwan, Abhinav Valada, and Wolfram Burgard.
\newblock Vlocnet++: Deep multitask learning for semantic visual localization
  and odometry.
\newblock {\em IEEE Robotics and Automation Letters}, 3(4):4407--4414, 2018.

\bibitem{sarlin2019coarse}
Paul-Edouard Sarlin, Cesar Cadena, Roland Siegwart, and Marcin Dymczyk.
\newblock From coarse to fine: Robust hierarchical localization at large scale.
\newblock In {\em Proceedings of the IEEE/CVF Conference on Computer Vision and
  Pattern Recognition}, pages 12716--12725, 2019.

\bibitem{sarlin2020superglue}
Paul-Edouard Sarlin, Daniel DeTone, Tomasz Malisiewicz, and Andrew Rabinovich.
\newblock Superglue: Learning feature matching with graph neural networks.
\newblock In {\em Proceedings of the IEEE/CVF conference on computer vision and
  pattern recognition}, pages 4938--4947, 2020.

\bibitem{sattler2016efficient}
Torsten Sattler, Bastian Leibe, and Leif Kobbelt.
\newblock Efficient \& effective prioritized matching for large-scale
  image-based localization.
\newblock {\em IEEE transactions on pattern analysis and machine intelligence},
  39(9):1744--1756, 2016.

\bibitem{sener2018multi}
Ozan Sener and Vladlen Koltun.
\newblock Multi-task learning as multi-objective optimization.
\newblock {\em Advances in neural information processing systems}, 31, 2018.

\bibitem{shavit2021learning}
Yoli Shavit, Ron Ferens, and Yosi Keller.
\newblock Learning multi-scene absolute pose regression with transformers.
\newblock In {\em Proceedings of the IEEE/CVF International Conference on
  Computer Vision}, pages 2733--2742, 2021.

\bibitem{shotton2013scene}
Jamie Shotton, Ben Glocker, Christopher Zach, Shahram Izadi, Antonio Criminisi,
  and Andrew Fitzgibbon.
\newblock Scene coordinate regression forests for camera relocalization in
  rgb-d images.
\newblock In {\em Proceedings of the IEEE conference on computer vision and
  pattern recognition}, pages 2930--2937, 2013.

\bibitem{sinha2018gradient}
Ayan Sinha, Zhao Chen, Vijay Badrinarayanan, and Andrew Rabinovich.
\newblock Gradient adversarial training of neural networks.
\newblock {\em arXiv preprint arXiv:1806.08028}, 2018.

\bibitem{song2023vp}
Ziying Song, Haiyue Wei, Caiyan Jia, Yongchao Xia, Xiaokun Li, and Chao Zhang.
\newblock Vp-net: Voxels as points for 3d object detection.
\newblock {\em IEEE Transactions on Geoscience and Remote Sensing}, 2023.

\bibitem{sun2021loftr}
Jiaming Sun, Zehong Shen, Yuang Wang, Hujun Bao, and Xiaowei Zhou.
\newblock Loftr: Detector-free local feature matching with transformers.
\newblock In {\em Proceedings of the IEEE/CVF conference on computer vision and
  pattern recognition}, pages 8922--8931, 2021.

\bibitem{taira2018inloc}
Hajime Taira, Masatoshi Okutomi, Torsten Sattler, Mircea Cimpoi, Marc
  Pollefeys, Josef Sivic, Tomas Pajdla, and Akihiko Torii.
\newblock Inloc: Indoor visual localization with dense matching and view
  synthesis.
\newblock In {\em Proceedings of the IEEE Conference on Computer Vision and
  Pattern Recognition}, pages 7199--7209, 2018.

\bibitem{turkoglu2021visual}
Mehmet~Ozgur Turkoglu, Eric Brachmann, Konrad Schindler, Gabriel~J Brostow, and
  Aron Monszpart.
\newblock Visual camera re-localization using graph neural networks and
  relative pose supervision.
\newblock In {\em 2021 International Conference on 3D Vision (3DV)}, pages
  145--155. IEEE, 2021.

\bibitem{valentin2016learning}
Julien Valentin, Angela Dai, Matthias Nie{\ss}ner, Pushmeet Kohli, Philip Torr,
  Shahram Izadi, and Cem Keskin.
\newblock Learning to navigate the energy landscape.
\newblock In {\em 2016 Fourth International Conference on 3D Vision (3DV)},
  pages 323--332. IEEE, 2016.

\bibitem{walch2017image}
Florian Walch, Caner Hazirbas, Laura Leal-Taixe, Torsten Sattler, Sebastian
  Hilsenbeck, and Daniel Cremers.
\newblock Image-based localization using lstms for structured feature
  correlation.
\newblock In {\em Proceedings of the IEEE International Conference on Computer
  Vision}, pages 627--637, 2017.

\bibitem{wallingford2022task}
Matthew Wallingford, Hao Li, Alessandro Achille, Avinash Ravichandran, Charless
  Fowlkes, Rahul Bhotika, and Stefano Soatto.
\newblock Task adaptive parameter sharing for multi-task learning.
\newblock In {\em Proceedings of the IEEE/CVF Conference on Computer Vision and
  Pattern Recognition}, pages 7561--7570, 2022.

\bibitem{wang2023sat}
Li Wang, Ziying Song, Xinyu Zhang, Chenfei Wang, Guoxin Zhang, Lei Zhu, Jun Li,
  and Huaping Liu.
\newblock Sat-gcn: Self-attention graph convolutional network-based 3d object
  detection for autonomous driving.
\newblock {\em Knowledge-Based Systems}, 259:110080, 2023.

\bibitem{wang2023multi}
Li Wang, Xinyu Zhang, Ziying Song, Jiangfeng Bi, Guoxin Zhang, Haiyue Wei,
  Liyao Tang, Lei Yang, Jun Li, Caiyan Jia, et~al.
\newblock Multi-modal 3d object detection in autonomous driving: A survey and
  taxonomy.
\newblock {\em IEEE Transactions on Intelligent Vehicles}, 2023.

\bibitem{wang2022matchformer}
Qing Wang, Jiaming Zhang, Kailun Yang, Kunyu Peng, and Rainer Stiefelhagen.
\newblock Matchformer: Interleaving attention in transformers for feature
  matching.
\newblock {\em arXiv preprint arXiv:2203.09645}, 2022.

\bibitem{wang2021continual}
Shuzhe Wang, Zakaria Laskar, Iaroslav Melekhov, Xiaotian Li, and Juho Kannala.
\newblock Continual learning for image-based camera localization.
\newblock In {\em Proceedings of the IEEE/CVF International Conference on
  Computer Vision}, pages 3252--3262, 2021.

\bibitem{wang2023mdl}
Shiguang Wang, Tao Xie, Jian Cheng, Xingcheng Zhang, and Haijun Liu.
\newblock Mdl-nas: A joint multi-domain learning framework for vision
  transformer.
\newblock In {\em Proceedings of the IEEE/CVF Conference on Computer Vision and
  Pattern Recognition}, pages 20094--20104, 2023.

\bibitem{xie2022deep}
Tao Xie, Kun Dai, Ke Wang, Ruifeng Li, Jiahe Wang, Xinyue Tang, and Lijun Zhao.
\newblock A deep feature aggregation network for accurate indoor camera
  localization.
\newblock {\em IEEE Robotics and Automation Letters}, 7(2):3687--3694, 2022.

\bibitem{xie2023deepmatcher}
Tao Xie, Kun Dai, Ke Wang, Ruifeng Li, and Lijun Zhao.
\newblock Deepmatcher: A deep transformer-based network for robust and accurate
  local feature matching.
\newblock {\em arXiv preprint arXiv:2301.02993}, 2023.

\bibitem{xie2020visual}
Tao Xie, Ke Wang, Ruifeng Li, and Xinyue Tang.
\newblock Visual robot relocalization based on multi-task cnn and
  image-similarity strategy.
\newblock {\em Sensors}, 20(23):6943, 2020.

\bibitem{xie2023farp}
Tao Xie, Li Wang, Ke Wang, Ruifeng Li, Xinyu Zhang, Haoming Zhang, Linqi Yang,
  Huaping Liu, and Jun Li.
\newblock Farp-net: Local-global feature aggregation and relation-aware
  proposals for 3d object detection.
\newblock {\em IEEE Transactions on Multimedia}, 2023.

\bibitem{xie2023poly}
Tao Xie, Shiguang Wang, Ke Wang, Linqi Yang, Zhiqiang Jiang, Xingcheng Zhang,
  Kun Dai, Ruifeng Li, and Jian Cheng.
\newblock Poly-pc: A polyhedral network for multiple point cloud tasks at once.
\newblock In {\em Proceedings of the IEEE/CVF Conference on Computer Vision and
  Pattern Recognition}, pages 1233--1243, 2023.

\bibitem{yang2022scenesqueezer}
Luwei Yang, Rakesh Shrestha, Wenbo Li, Shuaicheng Liu, Guofeng Zhang, Zhaopeng
  Cui, and Ping Tan.
\newblock Scenesqueezer: Learning to compress scene for camera relocalization.
\newblock In {\em Proceedings of the IEEE/CVF Conference on Computer Vision and
  Pattern Recognition}, pages 8259--8268, 2022.

\bibitem{yu2020learning}
Hailin Yu, Weicai Ye, Youji Feng, Hujun Bao, and Guofeng Zhang.
\newblock Learning bipartite graph matching for robust visual localization.
\newblock In {\em 2020 IEEE International Symposium on Mixed and Augmented
  Reality (ISMAR)}, pages 146--155. IEEE, 2020.

\bibitem{zhang2015visual}
Ji Zhang and Sanjiv Singh.
\newblock Visual-lidar odometry and mapping: Low-drift, robust, and fast.
\newblock In {\em 2015 IEEE International Conference on Robotics and Automation
  (ICRA)}, pages 2174--2181. IEEE, 2015.

\bibitem{zhou2020kfnet}
Lei Zhou, Zixin Luo, Tianwei Shen, Jiahui Zhang, Mingmin Zhen, Yao Yao, Tian
  Fang, and Long Quan.
\newblock Kfnet: Learning temporal camera relocalization using kalman
  filtering.
\newblock In {\em Proceedings of the IEEE/CVF Conference on Computer Vision and
  Pattern Recognition}, pages 4919--4928, 2020.

\bibitem{zhou2021patch2pix}
Qunjie Zhou, Torsten Sattler, and Laura Leal-Taixe.
\newblock Patch2pix: Epipolar-guided pixel-level correspondences.
\newblock In {\em Proceedings of the IEEE/CVF conference on computer vision and
  pattern recognition}, pages 4669--4678, 2021.

\end{thebibliography}
}

\end{document}